\documentclass[preprint,12pt]{elsarticle}

\usepackage[utf8]{inputenc} 
\usepackage[T1]{fontenc}    
\usepackage{hyperref}       
\usepackage{url}            
\usepackage{booktabs}       
\usepackage{amsfonts}       
\usepackage{nicefrac}       
\usepackage{microtype}      
\usepackage{lipsum}		
\usepackage{graphicx}
\usepackage[numbers]{natbib} 
\usepackage{doi}
\usepackage[usenames,dvipsnames]{xcolor}
\usepackage{float}

\usepackage{listings}
\usepackage{subcaption}
\usepackage{amsmath}
\usepackage{wrapfig}
\usepackage{multirow}
\usepackage{ulem}
\DeclareUnicodeCharacter{202F}{\,} 

\journal{Computers and Chemical Engineering}

\begin{document}

\begin{frontmatter}

\title{Learning from flowsheets: A generative transformer model for autocompletion of flowsheets}

\author[inst1]{Gabriel Vogel}
\author[inst1]{Lukas Schulze Balhorn}
\author[inst1]{Artur M. Schweidtmann\corref{cor1}}

\affiliation[inst1]{organization={Delft University of Technology, Department of Chemical Engineering, Process Intelligence Research Group},
            addressline={Van der Maasweg 9}, 
            city={Delft},
            postcode={2629 HZ},
            country={Netherlands}}
\cortext[cor1]{a.schweidtmann@tudelft.nl}

\begin{abstract}
	We propose a novel method enabling autocompletion of chemical flowsheets. 
	This idea is inspired by the autocompletion of text. 
	We represent flowsheets as strings using the text-based SFILES~2.0 notation and learn the grammatical structure of the SFILES~2.0 language and common patterns in flowsheets using a transformer-based language model. 
	We pre-train our model on synthetically generated flowsheet topologies to learn the flowsheet language grammar. 
	Then, we fine-tune our model in a transfer learning step on real flowsheet topologies. 
	Finally, we use the trained model for causal language modeling to autocomplete flowsheets.
	Eventually, the proposed method can provide chemical engineers with recommendations during interactive flowsheet synthesis. 
	The results demonstrate a high potential of this approach for future AI-assisted process synthesis but also reveal the limitations at the present state and the next steps that need to be taken to deploy this technique in realistic flowsheet synthesis scenarios.
\end{abstract}

\begin{graphicalabstract}
\includegraphics[width=\textwidth]{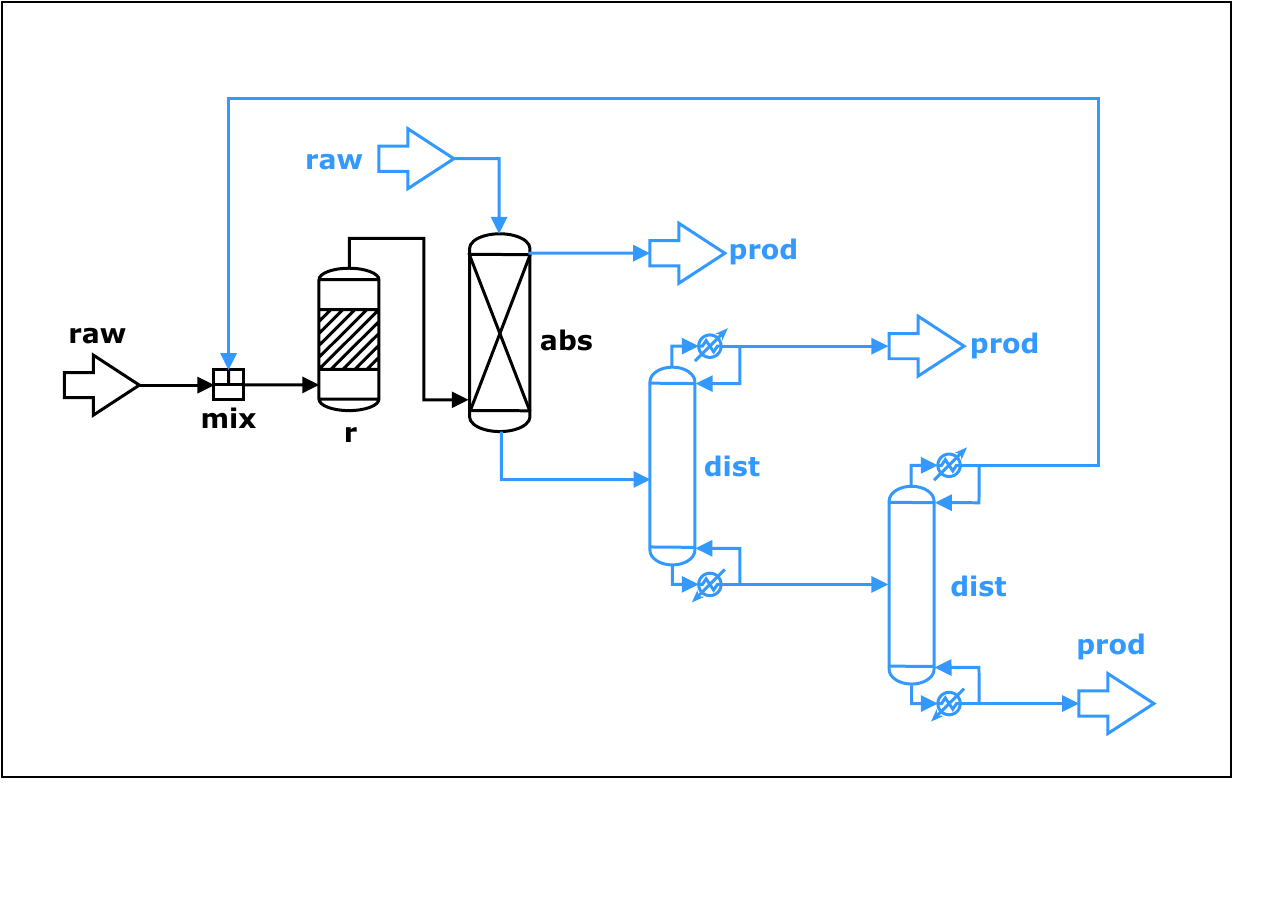}
\end{graphicalabstract}

\begin{highlights}
\item Design of transformer-based language model for autocompletion of flowsheets
\item Model is trained on flowsheet topologies represented as strings, in SFILES~2.0 format
\item Pre-training data set is generated synthetically
\item Data set for fine-tuning is derived from process simulation files
\item Model learns grammar of SFILES~2.0 language and completes process design patterns
\end{highlights}


\end{frontmatter}

\section{Introduction}
One important step in early process synthesis is the design of the flowsheet topology consisting of the selection and arrangement of unit operations and their connections. In an iterative procedure, engineers mostly perform this step manually based on experience, common design heuristics, and optimization-based methods~\cite{Chen2017, Biegler1997}.

Recent years have revealed a very high potential for artificial intelligence~(AI) in various disciplines of natural science and engineering (e.g.,~\cite{Jumper2021, Silver2016,Secinaro2021}). Advances in natural language processing~(NLP), a sub-field of AI, support several applications in daily life, such as intelligent text completion in search engines or email programs.
Recently, NLP methods have also been successfully applied to other domains. Github copilot uses text generation models to autocomplete computer code supporting software developers~\cite{Chen2021}. Another example are NLP applications in the molecular world using the text-based Simplified Molecule Input-Line Entry-System~(SMILES)~\cite{Weininger1989} notation which is a common representation format for molecules and chemical reactions. Recurrent neural networks~(RNNs) trained on SMILES strings have been used as generative models in de novo drug design~\cite{Segler2017}. Recently, transformer-based language models~\cite{Vaswani2017} have been applied to several tasks, such as the prediction of molecular properties~\cite{Chithrananda2020}, chemical reaction prediction with the Molecular Transformer~\cite{Schwaller2019}, and the prediction of retrosynthetic pathways~\cite{Schwaller2020}.

To this date, there have been only few previous works that use AI for process synthesis~\cite{Schweidtmann2021,Lee2018,Venkatasubramanian2018}. 
Notably, there have been many previous works on surrogate modeling and (superstructure) optimization~\cite{henao2011surrogate,kim2020surrogate,schweidtmann2019deterministic,huster2020deterministic} while only a few works actually create process topologies. 
Recently, there have been advances toward using Reinforcement Learning~(RL) approaches for process synthesis~\cite{Khan2020, Goettl2021, Goettl2021a, Midgley2020, Stops2022}. RL does not utilize data from existing flowsheets but strives to train an agent to learn the task of process synthesis. Contrary, the variety and large amount of historical chemical processes and their corresponding flowsheets pose many opportunities for data-driven models in the field of process synthesis~\cite{Zhang2018,Zheng2022}. However, unleashing the potential of data-driven AI in the chemical engineering domain also comes with several difficulties~\cite{Schweidtmann2021, Weber2021}, such as the design of meaningful information representations allowing advanced AI methods to understand complex data. 
In parallel to the developments of our proposed work, Oeing et al.~\cite{Oeing2022} very recently published a data-driven approach for AI-assisted development of piping and instrumentation diagrams~(P\&IDs). The authors collected 35 P\&IDs in DEXPI format and converted them to graphs. Then, they generated a training data set of linear sequences of six sequential unit operations by sampling random walks from the graphs. Oeing et al.~\cite{Oeing2022} trained an RNN on the sequential data to predict subsequent equipment. Furthermore, they used the graph data set to train a graph neural network~(GNN) to perform consistency checks during the drawing of P\&IDs, by comparing newly drawn equipment with patterns in the data set. 

We propose a novel methodology enabling autocompletion of chemical process flowsheets by learning the structure and patterns from a dataset of process flow diagrams~(PFDs) using a transformer-based NLP model. 
The underlying idea of our approach is to apply causal language modeling to autocomplete flowsheets comparable to sentence completion or, in general, text generation of human language. 
We mine and digitize PFDs from literature and also generate an additional synthetic PFD dataset. 
In order to train and deploy the transformer model on complex flowsheet topologies, we represent PFDs using our recent extension of the text-based Simplified Flowsheet Input-Line Entry-System~(SFILES) 2.0~\cite{d'Anterroches2006,GabrielVogel}. Eventually, the proposed flowsheet-completion approach aims to provide engineers with recommendations for the flowsheet structure during AI-assisted iterative process synthesis. 
In comparison to the work by Oeing et al.~\cite{Oeing2022}, we train a transformer model that learns from whole flowsheets rather than training an RNN on linear sequences. 
In particular, the SFILES~2.0 format allows us to represent and learn from complex flowsheet topologies including recycles and branchings (e.g., in distillation columns) while the previous work is limited to linear sequences. 
In addition, our model has a maximum input length of 512 unit operations per flowsheet compared to 6 units in the previous work. 
Likewise, our approach also autocompletes complete flowsheets (incl. complex recycles and branching) compared to the prediction of the next unit operation. Finally, we train our model on a comprehensive dataset of over 8,000 PFDs compared to 35 P\&IDs.

The remainder of this paper is structured as follows: In Section~\ref{sec:background}, we briefly explain the transformer architecture and the used flowsheet representations. Section~\ref{sec:GFT} explains all components of the flowsheet completion model, called Generative Flowsheet Transformer. Afterward, the used data for pre-training and fine-tuning of the model as well as the training procedure and results are described in Sections~\ref{sec:data} and~\ref{sec:results_training}. In Section~\ref{sec:results_completion}, we demonstrate the autocompletion technique using illustrative examples. In Section~\ref{sec:limitations} we also point out the current limitations of the proposed model regarding applicability in a context of realistic design problems and discuss the next essential steps that need to be taken to develop a more intelligent model.

\section{Background}\label{sec:background}
This section briefly explains the concepts of the original transformer model architecture~\citep{Vaswani2017} and further the typical modified model architecture for text generation applications. Afterward, it recaps the used flowsheet representations, namely flowsheet graphs and the SFILES~2.0 notation. The latter is used to represent the flowsheet data in a text-based manner and to use NLP models on the flowsheet data. 

\subsection{Original transformer architecture}
For many years, the state-of-the-art in the field of NLP had been recurrent architectures using LSTMs~\citep{Hochreiter1997} or GRUs~\citep{Cho2014}. In the recent past, Vaswani et al.~\citep{Vaswani2017} published the novel transformer model architecture using self-attention which is currently facing a hype in the AI community. It does not use recurrence but processes the entire sequence at once and therefore allows for significantly more parallelization in comparison to RNNs. Furthermore, self-attention layers are faster than recurrent layers in terms of computational complexity~\citep{Vaswani2017}. The reduced computational cost is one major reason for the growing popularity not only the field of NLP but also in the whole AI community. Transformer models have demonstrated breakthrough performances in a variety of applications~\citep{Vaswani2017, Schwaller2019, Lakew2018} and have, in many cases, the potential to replace RNNs and convolutional neural networks.

\begin{figure}[h!]
	\centering
	\includegraphics[scale=0.7]{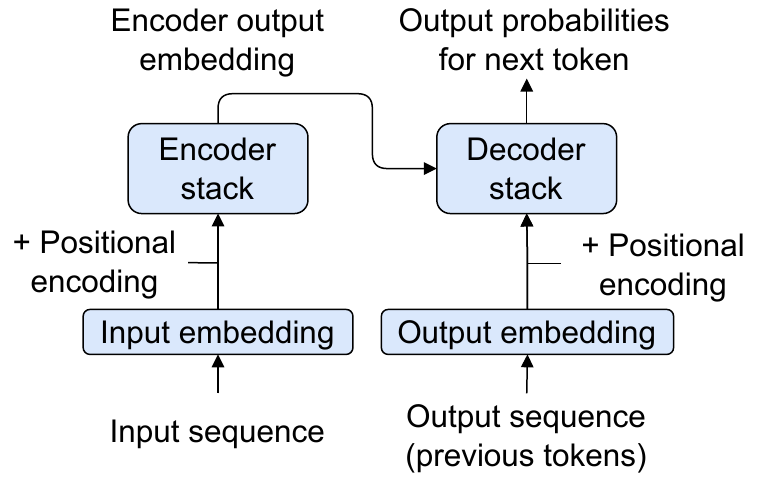}
	\caption{Simplified illustration of transformer architecture derived from~\citep{Vaswani2017} consisting of encoder and decoder stack.}
	\label{fig:Transformer_org_architecture}
\end{figure}

The original transformer architecture~\citep{Vaswani2017} is a neural sequence translation model which is usually used to translate an input sequence to an output sequence, e.g. a German to an English sentence. It consists of an encoder stack and a decoder stack, as shown in the simplified illustration in Figure~\ref{fig:Transformer_org_architecture}. The encoder converts a sequence of words or characters (input tokens) into a numerical representation that can be further processed in the model. It captures each token and its meaning in the sequence. This numerical representation is then fed to the decoder stack, which converts it back to a sequence of words or characters (generated output tokens) in the target language. The generation of the translated sequence is a token by token generation. This means, that the decoder repeatedly uses the encoder’s numerical output and previously generated output tokens to compute the probabilities for the next output token.

In the original architecture, the encoder stack and decoder stack each comprise $N=6$ identical layers. Multiple layers are used to enhance the model's ability to capture the whole complexity of a language. Each encoder layer contains two sub-layers with subsequent layer normalization. Each decoder layer contains three sub-layers with subsequent layer normalization. There are two general types of sub-layers used. First, the attention sub-layers comprise mathematical operations that help understand the language and the interdependence of tokens in a language. Second, the position-wise feed-forward sub-layers further process the attention output before feeding it to the next encoder or decoder layer.

Since recurrent components are completely removed in the transformer architecture, before input and output embeddings are passed to the encoder and decoder, respectively, positional encoding is applied (see Figure~\ref{fig:Transformer_org_architecture}). Positional encoding ensures that the information of the order of tokens in the sequence is taken into account. In the original implementation~\citep{Vaswani2017}, sine and cosine functions of different frequencies depending on the token position are added to the embeddings allowing the model to know each token position in the sequence. 

One important core component of the transformer architecture are the attention sub-layers. The calculation of attention takes a query vector \textbf{q}, key vector \textbf{k}, and value vector \textbf{v} for each input token and compares all queries against all keys resulting in scores for query-key compatibility. The compatibility scores are then used as weights to calculate the attention output as a weighted sum of the values. In practice, the attention is computed for all inputs of an input sequence in parallel, putting together all query, key, and value vectors in the query matrix $Q$, key matrix $K$, and value matrix $V$. This finally yields a matrix as attention output. In the transformer~\cite{Vaswani2017}, the scaled dot-product attention is implemented, which is calculated from the queries $Q$, keys $K$, and values $V$ as
\begin{equation}
    \mathrm{Attention}(Q,K,V)=\mathrm{softmax}\left(\frac{QK^T}{\sqrt{d_k}}\right)V,\label{eq:attention}
\end{equation}
where $d_k$ is the dimension of the key vectors. The transformer in specific uses multi-head attention, consisting of several scaled dot-product attention layers in parallel. For the multi-head attention, the keys, values, and queries are linearly projected into $h$ learned projections of the dimension $n_{embd}/h$, enabling multiple heads to learn from different representation subspaces in parallel~\citep{Vaswani2017}.

In the original architecture, multi-head attention is used as self-attention layers in the encoder, masked self-attention in the decoder, and encoder-decoder attention to combine the vector embedding of the encoder with the previous decoder outputs. Hereby, self-attention means that query, key, and value matrices are calculated from the same input sequence. Therefore, the computed self-attention represents each token and its meaning in the sequence. Self-attention in the encoder considers both the left and right context of each token~(bidirectional). Contrary, in the case of masked self-attention in the decoder, only the left context is used, meaning that subsequent positions of each token are masked out~(unidirectional). 
In practice, a mask is added to the matrix multiplication product of $Q$ and $K$ in Equation~\eqref{eq:attention}, i.e., by setting the softmax input of the respective positions to -$\infty$ and thus the respective outputs scores of the softmax to zero. This way the auto-regressive property of the decoder is preserved~\citep{Vaswani2017}. The encoder-decoder attention uses the encoder's output as keys and values and the decoder's masked self-attention output as queries. 

\subsection{Auto-regressive transformer for text generation}\label{sec:autoregressive_transformer}
\begin{figure}[h!]
	\centering
	\includegraphics[scale=0.7]{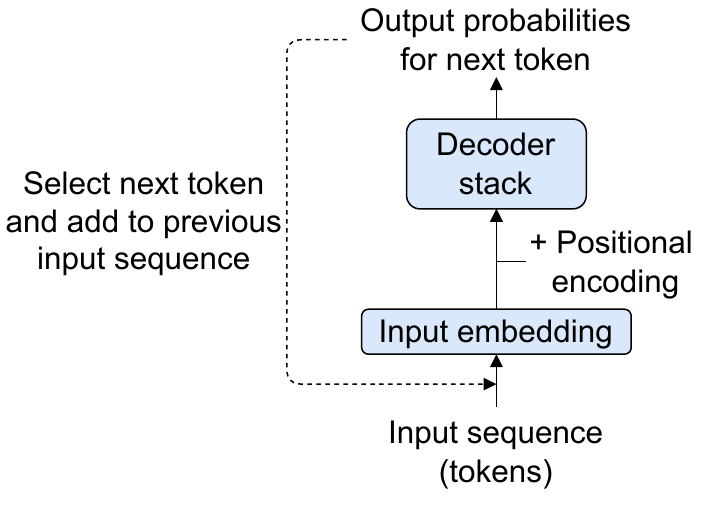}
	\caption{Simplified illustration of a decoder-only architecture for auto-regressive text-generation}
	\label{fig:decoder-only-transformer}
\end{figure}

Besides the original architecture, depending on the application, encoder and decoder stacks can be modified or left out entirely. Auto-regressive models for text generation, also called causal language modeling, typically only use the decoder part of the original transformer, as shown in Figure~\ref{fig:decoder-only-transformer}. Auto-regressive means that previously generated tokens are added to the input sequence for the next token generation. More specifically, at each time step, the decoder-only transformer model outputs probabilities for different tokens suitable as the next token in the sequence. The selected token is then added to the previous input sequence (dashed line in Figure~\ref{fig:decoder-only-transformer}) before the decoder computes the following outputs. 

In recent years, many model architectures for text generation and pre-trained models were published, such as GPT-2~\citep{Radford2019}, Transformer-XL~\citep{Dai2019}, and XLNet~\citep{Yang2019}.

\subsection{Graph- and text-based representation of flowsheets}\label{sec:flowsheet_representations}
This section briefly summarizes the flowsheet representations that we used in the data processing steps. The data sets, described in Section \ref{sec:data}, are first created in graph format and later converted to the text-based SFILES~2.0 strings~\cite{d'Anterroches2006,GabrielVogel}. 

\begin{figure}[h!]
	\centering
	\includegraphics[scale=0.6]{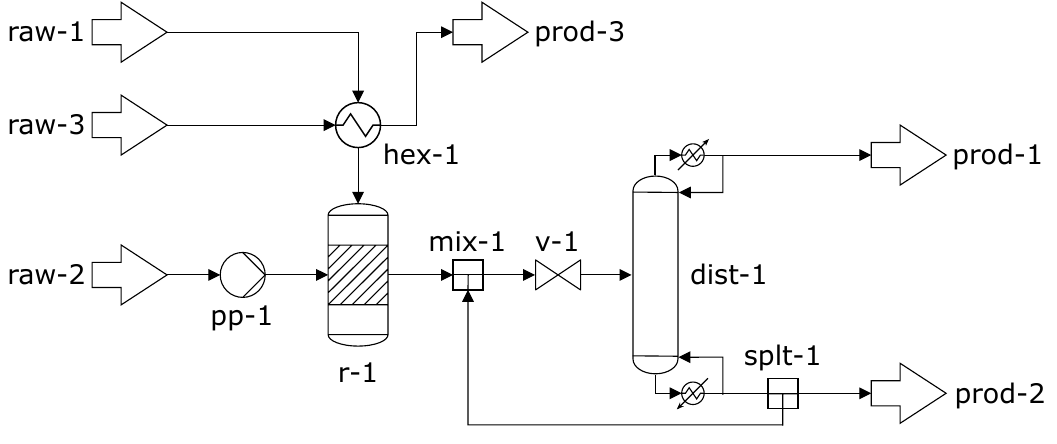}
	\caption{Simple chemical process flowsheet with branchings, recycle stream, and different mass trains}
	\label{fig:Flowsheet_intro}
\end{figure}

\begin{figure}[h!]
	\centering
	\includegraphics[scale=0.6]{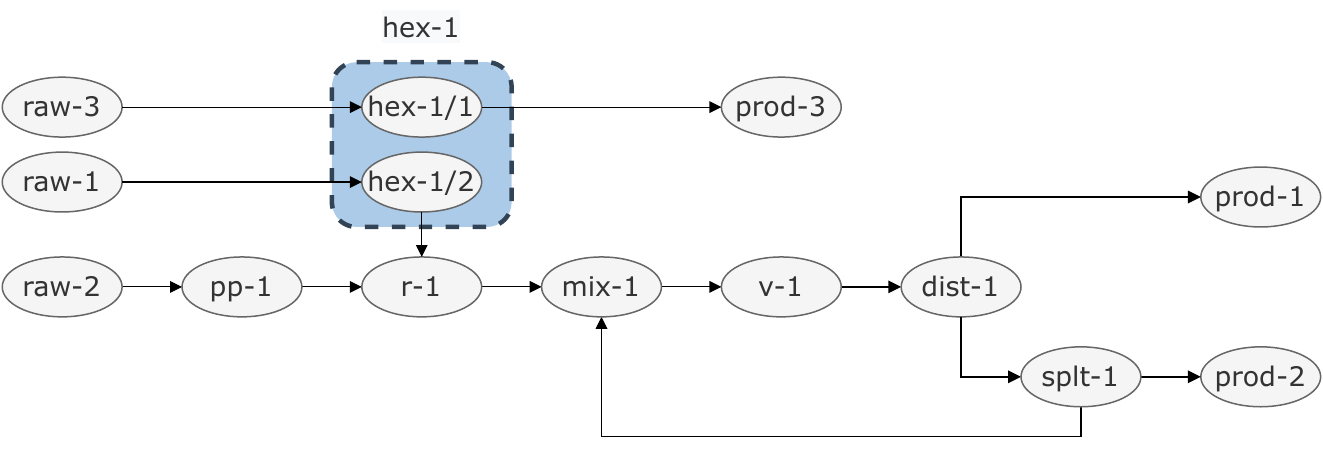}
	\caption{Graph representation of flowsheet in Figure~\ref{fig:Flowsheet_intro}}
	\label{fig:intro_pfd_graph}
\end{figure}

Flowsheets can be represented as directed heterogeneous graphs~\cite{GabrielVogel, Zhang2018} with unit operations as nodes and stream connections between the unit operations as directed edges. Figure~\ref{fig:Flowsheet_intro} shows an example flowsheet with a two-stream heat exchanger, reactor, distillation system, and a recycle loop. This flowsheet can be represented as the flowsheet graph shown in Figure~\ref{fig:intro_pfd_graph}, whereby the heat exchanger unit operation is divided into two nodes, one for each stream compartment.


Using the SFILES~2.0 notation~\cite{GabrielVogel} and our open-source conversion implementation, provided in our \href{https://github.com/process-intelligence-research/SFILES2}{GitHub repository}~\cite{Vogel2022} yields the corresponding SFILES~2.0 string
\begin{lstlisting}
(raw)(hex){1}(r)<&|(raw)(pp)&|(mix)<1(v)(dist)[{tout}(prod)]{bout}(splt)1(prod)n|(raw)(hex){1}(prod)
\end{lstlisting}
for the flowsheet graph in Figure~\ref{fig:intro_pfd_graph}. In an SFILES 2.0 string, unit operations are represented in parentheses. Two consecutive unit operations in parentheses imply a directed stream connection from the left to the right unit operation. In the case of branching in the process such as after a distillation column \texttt{(dist)} (one input and two output streams), the branches are noted in brackets. The brackets are omitted for the branch that is noted last in the SFILES~2.0 string (in the given example the branch continuing with \texttt{(splt)}). Recycles are noted using numbers \# to reference the recycle start node (here \texttt{(splt)}) and <\# to reference the recycle end node (here \texttt{(mix)}). Furthermore, tags in braces are used to indicate whether the branch is a top or bottom product. In the case of converging branches, the second branch is inserted in the string, surrounded by <\&| and \&| (here \texttt{<\&|(raw)(pp)\&|}). Multi-stream heat exchangers are separated in one node per stream compartment and marked with a number in braces, capturing which streams are heat-integrated.

\section{Generative Flowsheet Transformer}\label{sec:GFT}
This section first provides an overview of the general procedure from an incomplete to a completed flowsheet in Section~\ref{sec:overview}. Then, we specify details of the single components of the Generative Flowsheet Transformer, including the SFILES~2.0 tokenization (Section~\ref{sec:tokenization}), the decoder stack architecture (Section~\ref{sec:decoder-only-model}), and the decoding strategies for flowsheet completion (Section~\ref{sec:completion-strategy}).

\subsection{Overview}\label{sec:overview}
\begin{figure}[h!]
\centering
\begin{subfigure}[b]{\textwidth}
   \includegraphics[width=1\linewidth]{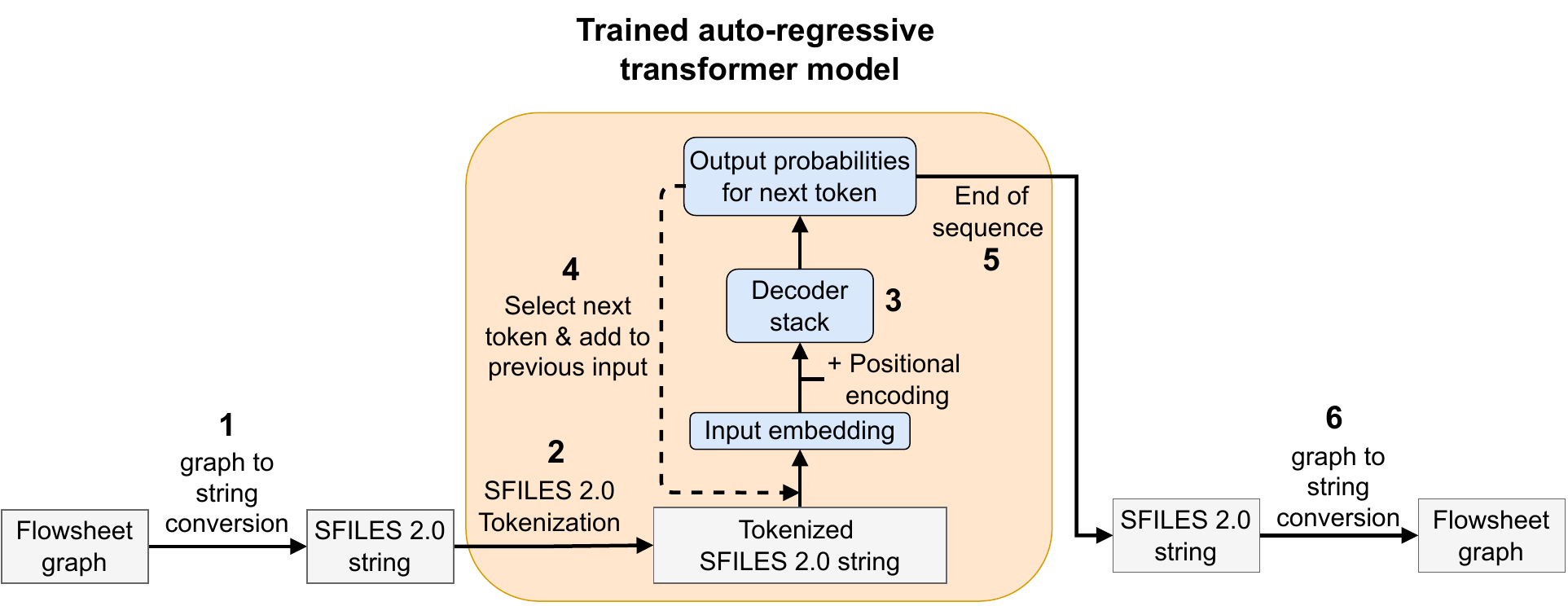}
   \caption{}
   \label{fig:Detailed_method} 
\end{subfigure}
\par\bigskip
\begin{subfigure}[b]{\textwidth}
   \includegraphics[width=1\linewidth]{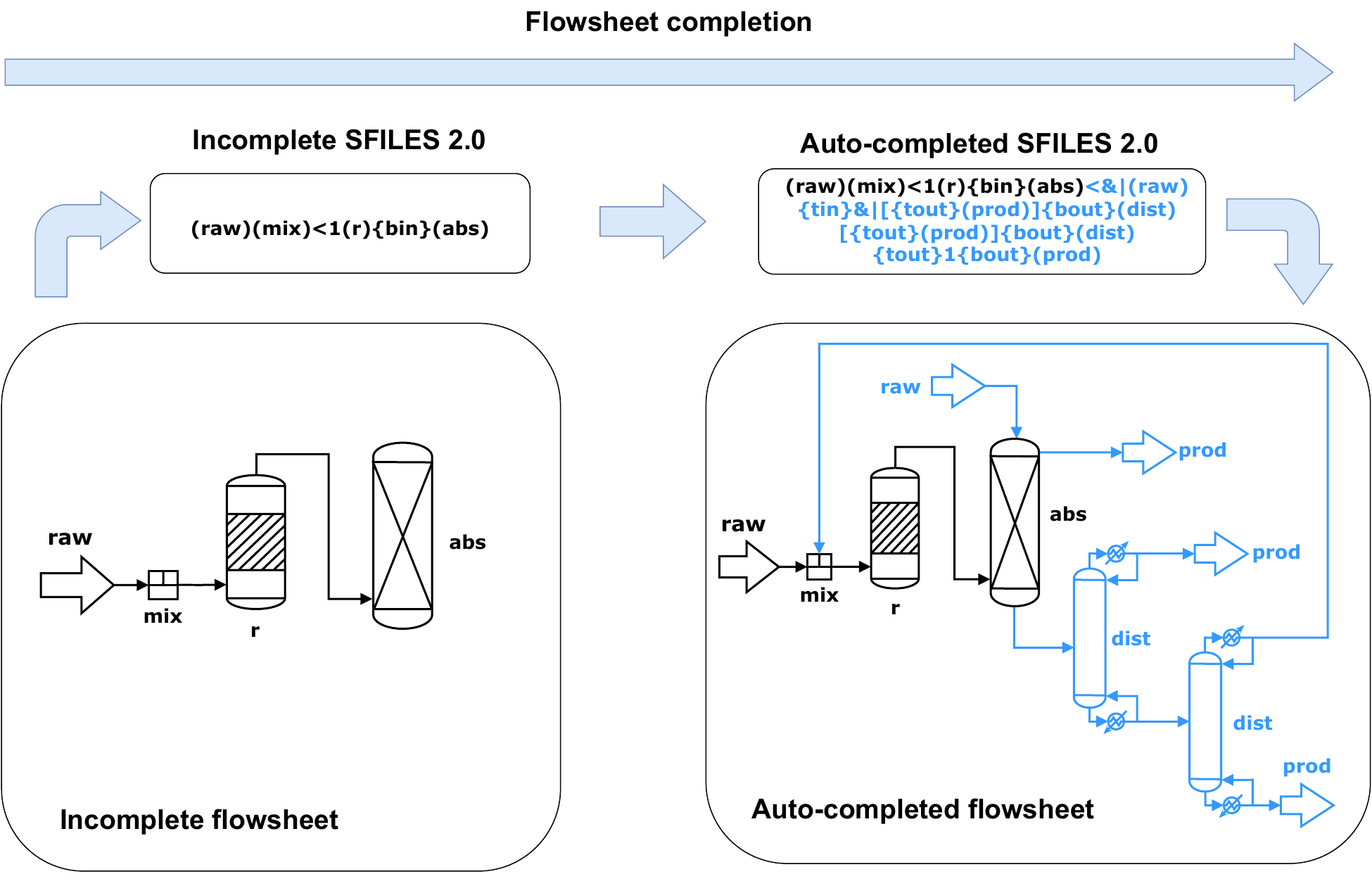}
   \caption{}
   \label{fig:Detailed_example}
\end{subfigure}

\caption{Overview of flowsheet completion with the Generative Flowsheet Transformer (a) Incomplete flowsheet graph is converted to incomplete SFILES~2.0 string (1). Auto-regressive transformer model completes string (2,3,4,5). autocompleted SFILES~2.0 string is converted to completed flowsheet graph (6). (b) Example autocompletion of a flowsheet}\label{fig:Detailed_overview}
\end{figure}

The proposed flowsheet completion methodology is illustrated in Figure~\ref{fig:Detailed_method} and described in the following. 
In step 1, the incomplete flowsheet graph is converted to the corresponding SFILES~2.0 string, as described in Section~\ref{sec:flowsheet_representations}. Afterward, in step 2, the string is tokenized using the SFILES~2.0 tokenizer (Section~\ref{sec:tokenization}). The resulting input embedding is passed to the decoder stack (Section~\ref{sec:decoder-only-model}), which computes the output probabilities for the next token prediction. Step 4 comprises the selection of the next token from the decoder output, which is determined by the decoding strategy, as described in Section~\ref{sec:completion-strategy}. After the SFILES~2.0 string completion is finished by reaching a defined end token or by controlling the number of tokens to be generated (step 5), the string is converted back to the autocompleted flowsheet graph (step 6).
This procedure could be embedded in process simulation software for the interactive autocompletion of flowsheets. 

\subsection{Tokenizer}\label{sec:tokenization}
Tokenization is the first text processing step in NLP. Therein, text input is usually processed as a sequence of tokens, whereby the tokens are either words or other chunks of the input sequence. Common strategies to identify tokens in natural language are word-level, WordPiece~\citep{Wu2016}, and most recently, Byte Pair Encoding~(BPE) tokenization~\citep{Sennrich2015}. After tokenizing the input sequence, each token is converted to a vector of size $n_{embd}$, called numerical embedding. This vector is calculated by multiplying a one-hot-encoded vector of the token with a weight matrix, learned by the language model during the training procedure. Combining the input vectors of a sequence of tokens yields a matrix that is called the input embedding. 

We trained our language model on the SFILES~2.0 language, which significantly differs from human language. Therefore, common tokenization strategies such as word-level, WordPiece, and Byte-pair encoding are not suitable for our task. Thus, we developed a tailored tokenizer for SFILES~2.0 which identifies, for instance, a unit operation (e.g. \texttt{(hex)}), a branching, stream tags (e.g. \texttt{\{tout\}}), a recycle connection number, etc., using a regular expression. This approach is inspired by the tokenizer for the molecular representation using SMILES~\citep{Schwaller2018}. Our proposed regular expression is 
\begin{lstlisting}[breaklines]
    r"(\(.*?\)|\{.*?\}|\%\([0-9]{3}\)|\%[0-9]{2}|\]|\[|\<.?[0-9]|\<\&\||(?<!\<)\&\||n\||(?<!\&)(?<!n)\||\&(?!\|)|\/[0-9]|[0-9])".
\end{lstlisting}
The following example shows an SFILES~2.0 string before and after the tokenization. 
\begin{lstlisting}
    Before:  (raw)(hex)(r)(mix)<1(v)(dist)[{tout}(prod)]{bout}(splt)1(prod)
\end{lstlisting}
\begin{lstlisting}
    After:  (raw)  (hex)   (r)  (mix)  <1  (v)  (dist)  [  {tout}   (prod)  ]  {bout}  (splt)  1  (prod)
\end{lstlisting}

\subsection{Decoder-only architecture for causal language modeling}\label{sec:decoder-only-model}
Since we aim to complete SFILES~2.0 strings using causal language modeling, our proposed architecture is based on an auto-regressive decoder-only transformer model as described in Section~\ref{sec:autoregressive_transformer}. 
The SFILES~2.0 language with a small vocabulary size of less than 100 is rather simple compared to human language with a vocabulary size of approximately 500k words in Finnish or approximately 90k words in English~\cite{birch2008predicting}. Thus, we use the architecture of the small version of the GPT-2 transformer model~\cite{Radford2019}, which contains a decoder stack consisting of $N=12$ decoder layers. Each decoder layer comprises a masked multi-head self-attention sub-layer with $h=12$ attention heads and a feed-forward sub-layer. With an embedding dimension of $n_{embd}=768$ the number of parameters in this architecture adds up to 85.9 million. The high number of parameters increases the capacity of the model to capture the entire complexity of a language.
On the other hand, this also increases the risk of overfitting the model on our training data. The GPT-2 configuration we used, by default implements dropout during the training procedure in the embedding, attention, and fully-connected layers to counteract overfitting. Furthermore, we make use of a technique called early-stopping during the training procedure, which monitors the error on a validation set and stops the training once this error does not decrease anymore.
In future works, a more rigorous investigation of the optimal model size and hyperparameters could be performed using Bayesian optimization or information criteria.

\subsection{Decoding strategies}\label{sec:completion-strategy}
When completing a flowsheet, as shown in Figure~\ref{fig:Detailed_overview}, the input is either an empty flowsheet represented as an empty input sequence or an incomplete flowsheet represented as an incomplete SFILES~2.0 string. In this work, we use open-ended text generation to complete SFILES~2.0 strings. Starting from an input sequence of $m$ tokens $t_1,..,t_m$, considered the context for the language generation, the trained model generates $n$ tokens in an auto-regressive manner until the sequence is completed~\citep{Holtzman2020} (see Section~\ref{sec:decoder-only-model}). With the assumption that the probability distribution is composed of the product of conditional next token distributions~\citep{Holtzman2020}, the final probability of the sequence is given as
\begin{equation}
    P(t_{1:m+n})=\prod_{i=1}^{m+n} P(t_i|t_{1},...,t_{i-1})\label{eq:final_cond_prob}.
\end{equation}
Different decoding~(or text generation) strategies define the selection of the next token at each step. 
Common strategies are greedy search, beam search, top-$k$ sampling, and top-$p$ sampling. 

Greedy search simply selects the token with the highest probability at each time step. This increases the chance of missing high-probability tokens hidden after low-probability tokens. Therefore, greedy search often does not lead to sequences with a maximized final probability.

Beam search aims to overcome the weakness of greedy search, i.e. by finding the optimal sequence by maximizing the final probability $P(t_{1:m+n})$ according to Equation \eqref{eq:final_cond_prob}. Constructing a tree of the possible tokens at each time step, the beam search is similar to a breadth-first search through the tree to find the path~(sequence) with the highest likelihood. The only difference is that beam search takes a parameter, beam-width $w_{beam}$, which determines the number of considered tokens at each step. However, several studies found that beam search often yields repetitive text~\citep{Shao2017,Vijayakumar2016} and lower-quality text compared to other decoding strategies that include sampling~\citep{Fan2018,Holtzman2020}.

Top-$k$ and top-$p$ sampling increase the chance of also selecting lower-probability tokens which often yields more diversity of the generated sequences. The parameters $k$ and $p$ truncate the distribution of token probabilities and prevent the model from randomly sampling from the whole probability distribution but rather from a limited selection of more-likely tokens. 

The top-$p$ sampling decoding strategy addresses the problem of fixed $k$ values in top-$k$ sampling. Its key idea is to use the shape of the probability distribution to define a confidence region at each time step~\citep{Holtzman2020}. The number of candidates in this confidence region dynamically changes and thus eliminates the problem of selecting a fixed number $k$ of candidates, as in top-$k$ sampling. With the definition of a threshold $p$, the smallest set of top-$p$ candidates $C^{(p)}$ is selected, such that the cumulative probability mass exceeds the threshold. Equation~\eqref{eq:top-p-condition} shows this relation.
\begin{equation}
    \sum_{t\in C^{(p)}} P(t|t_{1:i-1})\geq p \label{eq:top-p-condition}
\end{equation}
Depending on the shape of the probability distribution, the number of candidates within that set changes. Usually, high values of $p$ (between 0.9 and 1) are selected such that the set $C^{(p)}$ will take up most of the probability mass representing the nucleus~\citep{Holtzman2020}. The probability distribution is then re-scaled only using the set $C^{(p)}$ and the next token is sampled from the new distribution~\citep{Holtzman2020}.

We use beam search with a fixed beam-width of $w_{beam}=5$. The selection of $w_{beam}$ is a compromise of getting better results for long generated sequences while keeping the computational cost low. To introduce more diversity in the generated sequences, we choose top-$p$ sampling as our second decoding strategy with a cumulative probability threshold of $p=0.9$. 

\section{Data}
\label{sec:data}
This section explains the data sets that we use for the training of the Generative Flowsheet Transformer. The pre-training data set comprises synthetically generated flowsheets, while the fine-tuning data set consists of real flowsheets extracted from chemical process simulation files.

\subsection{Generated data for pre-training}\label{sec:artificial_data}

Language models are complex architectures with a high number of trainable weights. When dealing with human language, training these weights usually requires several Gigabytes to Terrabytes of text~\cite{Radford2019,Brown2020}. The SFILES~2.0 language is less complex than human language, consisting of a small vocabulary. Still, for the model to learn the SFILES~2.0 grammar of flowsheet topologies, we need a reasonable amount of data. There is currently no (public) database or data set of flowsheets~\cite{Schweidtmann2022Flowsheetmining}. Hence, we construct a synthetic data set of realistic flowsheet graphs for pre-training of our model. 

\begin{figure}[h!]
	\centering
	\includegraphics[width=\textwidth]{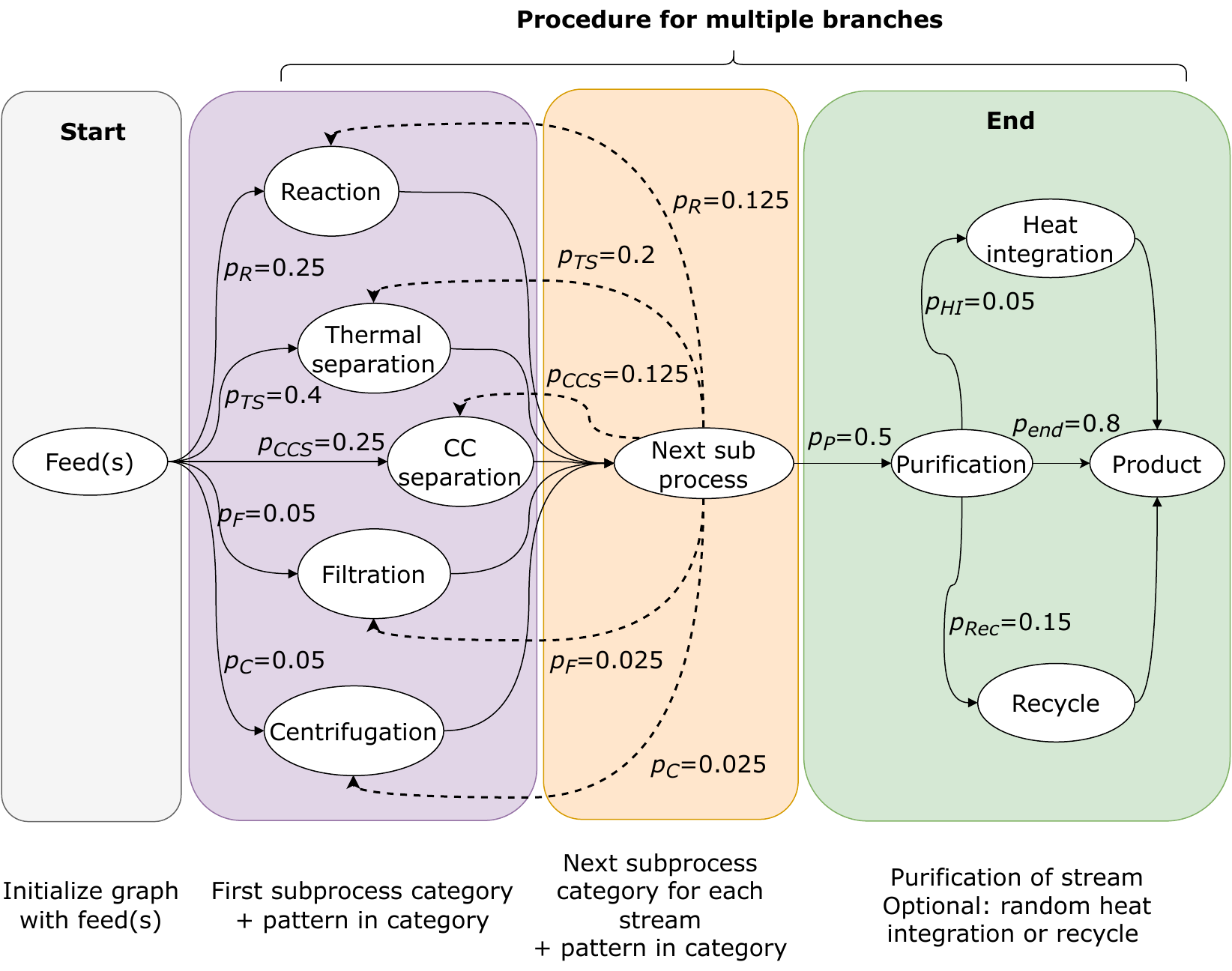}
	\caption{Random flowsheet graph generation scheme}
	\label{fig:Random-flowsheet-gen}
\end{figure}

The synthetic flowsheet generation builds up flowsheet graphs by a drawing random samples from a Markov chain-like process based on known flowsheet design heuristics.
We subdivide flowsheets into the following sub-process categories: 
Initialization with feed(s), reaction, thermal separation (distillation, rectification), countercurrent separation (absorption, extraction), filtration (gas,liquid), centrifugation, and purification as the last subprocess.
As illustrated in Figure~\ref{fig:Random-flowsheet-gen}, after initializing the flowsheet graph with raw materials, including feed pre-processing, the selection of the first sub-process, excluding purification, is a Markov transition with fixed probabilities where transition probabilities do not depend on previous unit operations. Within each sub-process, we further sample from a set of patterns specifying how the inlet and outlet streams are processed, e.g., with additional temperature or pressure change unit operations. Also, we include design heuristics such as adding recycles, performing heat integration in the reaction sub-process, or adding reactants~\citep{Zhang2018}. 
In general, the sub-processes lead to several outlet streams, in the following referred to as branches.
For each branch, we transition to the "Next sub-process" state, followed by a Markov transition to the next sub-process. This selection differs from the first sub-process selection by the additional purification sub-process. Note that once a branch reaches the purification step, it is determined to end as a product. After each branch ended in the purification step, the flowsheet graph generation is complete. Table~\ref{tab:data-props} summarizes the properties of the generated data set after removing duplicates and flowsheet graphs with more than 50 nodes. The SFILES~2.0-based data set for pre-training is obtained by automatic conversion of the generated graphs to SFILES~2.0 strings~\cite{GabrielVogel}.
\subsection{Real flowsheet data for fine-tuning}\label{sec:real_data}
We collected 223 Aspen and DWSIM chemical process simulation files from the public domain. We automatically extract information from the simulation files and convert them to flowsheet graphs using our in-house software. Then, the process graphs are converted to SFILES~2.0 strings using our open-source code~\cite{GabrielVogel}. Finally, we use the extracted flowsheet data for fine-tuning the pre-trained model. 

Table~\ref{tab:data-props} summarizes some key statistics of the synthetic and real flowsheet data set. The real data set shows a higher average node number, standard deviation, and vocabulary size compared to the generated data. 

\begin{table}[h!]
	\caption{Properties of the used data sets and the number of samples in training~(tr), validation~(val), and test~(te) data set  }
	\centering
	\begin{tabular}{r|rr}
		\toprule
		& Generated data set &Real data set\\
		\midrule
		$\mathrm{samples}_{tr}$&6,362&178\\
		$\mathrm{samples}_{val}$&796&23\\
		$\mathrm{samples}_{te}$& 795&22\\
		$\overline{nr_{nodes}}$&19& 25\\
		$\sigma(nr_{nodes})$&10&41\\
		vocabulary size&53&89\\
		\bottomrule
	\end{tabular}
	\label{tab:data-props}
\end{table}

\section{Training results and discussion}\label{sec:results_training}
The following section first describes and discusses the results of the transformer model training, including pre-training and fine-tuning. Furthermore, the model is evaluated based on perplexity, a metric commonly used in NLP. 

\subsection{Model training}\label{sec:training}
For pre-training and fine-tuning, we divide the dataset into a training~(80\%), validation~(10\%), and test set~(10\%). Table~\ref{tab:data-props} shows the training, validation, and test set sizes. During pre-training, we used a batch size of 8, while during fine-tuning on the small data set, we used a batch size of 2 to update weights more frequently. The validation set was evaluated every 200 steps during pre-training and every 20 steps during fine-tuning. In both training procedures, we use early stopping with to prevent overfitting of the model to the training set, with the early stopping patience set to 3 during pre-training and to 20 during fine-tuning.

\begin{figure}[h!]
\centering
\begin{subfigure}{.5\textwidth}
  \centering
  \includegraphics[width=.9\linewidth]{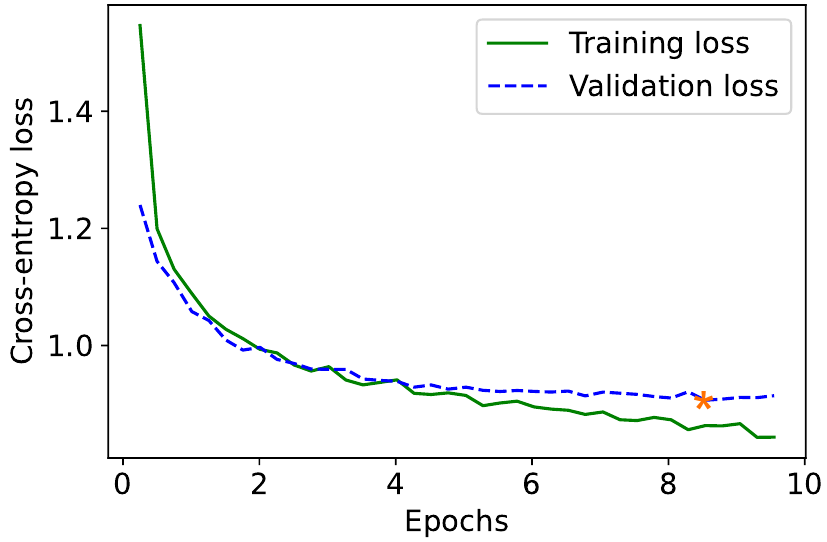}
  \caption{Training and validation loss during pre-training}
  \label{fig:pre-train-loss}
\end{subfigure}%
\begin{subfigure}{.5\textwidth}
  \centering
  \includegraphics[width=.9\linewidth]{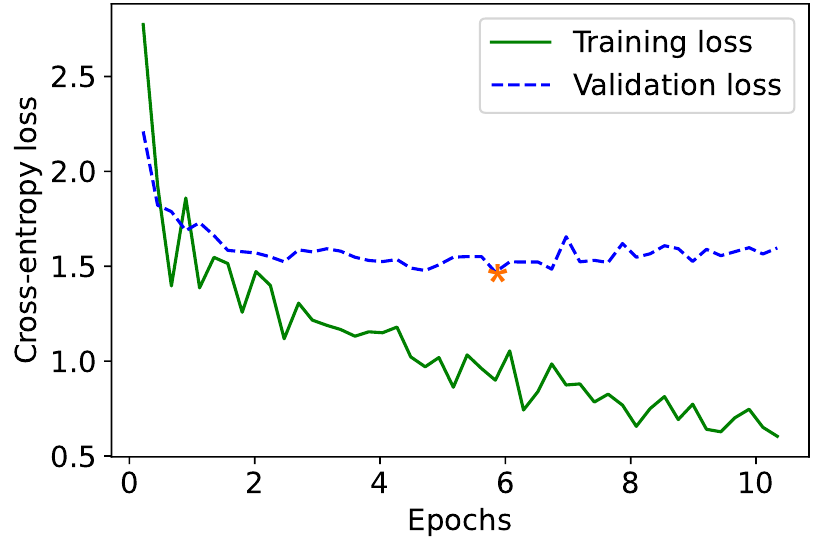}
  \caption{Training and validation loss during fine-tuning}
  \label{fig:fine-tune-loss}
\end{subfigure}
\caption{Training and validation cross-entropy loss over number of training epochs. The asterisk marks the lowest validation loss during training which is used in the early stopping strategy to load the model parameters.}
\label{fig:loss-curves}
\end{figure}

Figure~\ref{fig:pre-train-loss} shows the loss curves of the pre-training. The pre-training results show only small fluctuations in the loss curves and the gap between train and validation loss is small. This demonstrates a small generalization error on the synthetic dataset. 
This is likely due to the limited variability in the pre-training data, which is generated synthetically. Essentially, the data sets are drawn from the same probability distribution. Thus, the learned flowsheet patterns from the training set are representative for those in the synthetic validation and test set.

Figure~\ref{fig:fine-tune-loss} shows the loss curves of the fine-tuning.
In contrast to the pre-training, the fine-tuning shows considerable fluctuations of the loss curves and a larger gap between train and validation loss.
The reason for this learning behaviour is likely caused by the real flowsheet data. 
The real data consist of fewer flowsheets with higher variations in the number of nodes per flowsheet and an extended vocabulary size, i.e., more unit operation categories, as shown in Table~\ref{tab:data-props}. This leads to heavier fluctuations and a a larger generalization error. In the future, we envision to create a public knowledge graph for flowsheets that will mitigate this issue~\cite{Schweidtmann2022Flowsheetmining}.

\subsection{Perplexities}
In the following, we evaluate the model results using the perplexity $PP$, which is the most common metric for the evaluation of text generation models. It measures the ability of a probability model to predict a sequence of unseen test data and is defined as the exponential of the negative average log-likelihood of a sequence, which is equivalent to the exponential of the cross-entropy loss obtained during the model training. For the sequence $T=(t_1,...,t_n)$ of $n$ tokens, we calculate the perplexity as
\begin{equation*}
    PP(T)=\exp\left(-\frac{1}{n}\sum_{i}^{n}\log P(t_i|t_{1:i-1})\right).
\end{equation*}
In this equation, $P(t_i|t_{1:i-1})$ is the predicted conditional probability for token $t_i$ given the tokens $t_1, ..., t_{i-1}$. Low perplexities indicate that the model does well in predicting the next tokens in the test data, while high perplexities imply that the learned probability distribution is different from the probability distribution of the test data.

\begin{table}[h!]
	\caption{Perplexities of pre-trained and fine-tuned model for all used data sets}
	\centering
	\begin{tabular}{llrrr}
		\toprule
    	Model& data set& $PP_{tr}$ & $PP_{val}$ & $PP_{te}$  \\
		\midrule
		\multirow{2}{*}{Pre-trained}&Generated data & 2.28& 2.48 &2.47\\
		&Real data  &41.61 &64.16&25.91\\
		\hline
		\multirow{2}{*}{Fine-tuned}&Generated data  & 5.04 & 5.27 & 5.33 \\
		&Real data  & 2.17 & 3.98&4.75\\
		\bottomrule
	\end{tabular}
	\label{tab:perplexities}
\end{table}

Table~\ref{tab:perplexities} shows the perplexities of the pre-trained and fine-tuned models using both data sets. 
The pre-trained model shows similar perplexities on the generated train-, validation-, and test-set, because the data is all drawn from the same probability distributions of unit operations and patterns. This is consistent with the small generalization error, observed in Section~\ref{sec:training}.

Pre-training with synthetic data is necessary but not sufficient for predicting real flowsheet data. Table~\ref{tab:perplexities} shows that the pre-trained model performs poorly on the data set of real flowsheets.
Thus, we conclude that the generated data does not represent real flowsheet topologies sufficiently and fine-tuning on real data is necessary.
Note that we also attempted to train a model from scratch only using the real data. Nevertheless, it resulted in worse training, validation, and test loss than our proposed approach. This is expected as transformer models are known to be very data hungry. We conclude that the pre-training on a larger corpus of SFILES~2.0 enables the model to learn the basic grammatical structure of the SFILES~2.0 notation. The learned knowledge enhances the model's performance when fine-tuning it on the real data.

The fine-tuning significantly improves the model performance on real data. 
After fine-tuning, the model performance on unseen data increases from 25.91 to 4.75. 
This is remarkable as the final performance on the real flowsheet data is in the same order of magnitude as the pre-trained model on the generated dataset (2.47). The results show that the model has the potential to learn flowsheet patterns from real flowsheets. Notably, the fine-tuned model performs slightly worse on the generated data (5.33) than the pre-trained model (2.47), which is expected as the model replaced some of the learned patterns of the generated data set with the patterns contained in real flowsheets.

\section{Illustrative examples and discussion}
\label{sec:results_completion}
In this section, we illustrate and compare the proposed autocompletion techniques on a few examples. We further discuss the applicability of the results and point out the limitations that our model faces at the present state.

\subsection{Example 1}\label{sec:completion1} 
\begin{figure}[H]
	\centering
	\begin{subfigure}{\textwidth}
      \centering
      \includegraphics[scale=0.5]{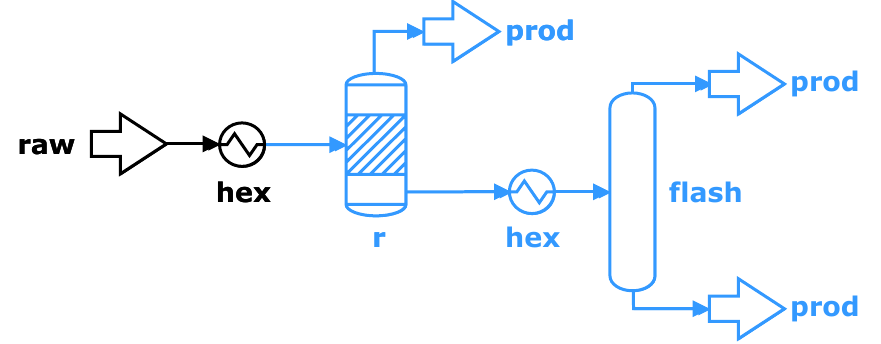}
      \caption{1. generated sequence}
      \label{fig:Ex1b1}
    \end{subfigure}
    \par\bigskip 
    \begin{subfigure}{\textwidth}
       \centering
       \includegraphics[scale=0.5]{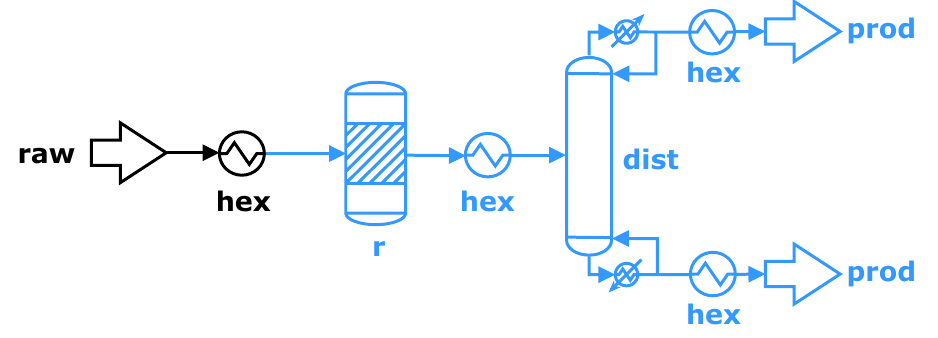}
       \caption{2. generated sequence}
       \label{fig:Ex1b2}
    \end{subfigure}
    \begin{subfigure}{\textwidth}
    \centering
       \includegraphics[scale=0.5]{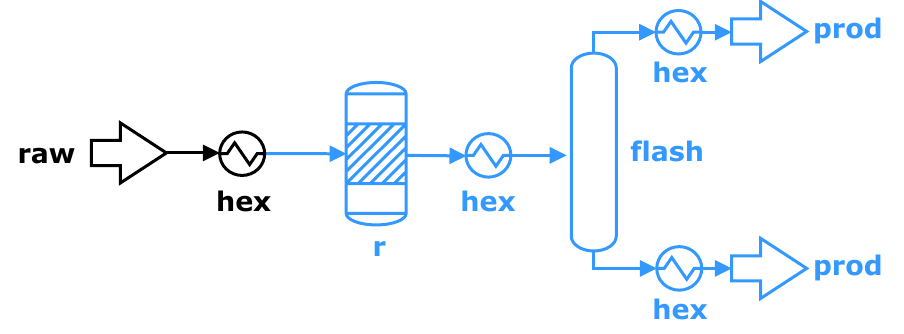}
       \caption{3. generated sequence}
       \label{fig:Ex1b3}
    \end{subfigure}
	\caption{Example 1: Completed flowsheets using beam search}
	\label{fig:BeamEx1complete}
\end{figure}

\begin{figure}[H]
\centering
\begin{subfigure}{\textwidth}
  \centering
  \includegraphics[scale=0.5]{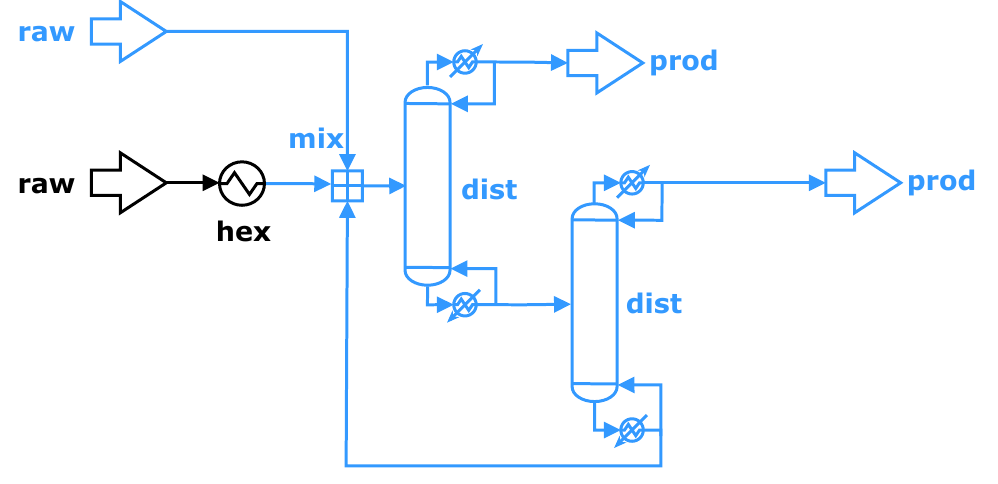}
  \caption{1. generated sequence}
  \label{fig:Ex1p1}
\end{subfigure}
\par\bigskip 
\begin{subfigure}{\textwidth}
   \centering
   \includegraphics[scale=0.5]{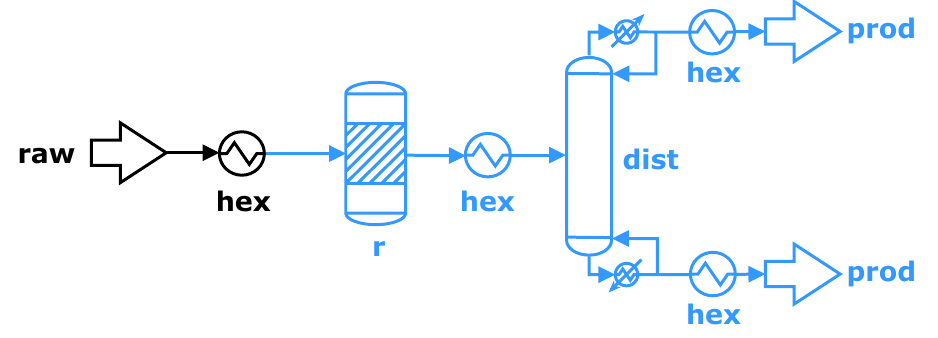}
   \caption{2. generated sequence}
   \label{fig:Ex1p2}
\end{subfigure}
\par\bigskip 
\begin{subfigure}{\textwidth}
\centering
   \includegraphics[scale=0.5]{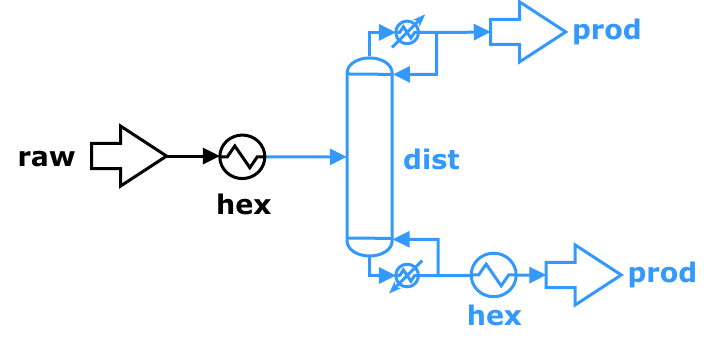}
   \caption{3. generated sequence}
   \label{fig:Ex1p3}
\end{subfigure}
\caption{Example 1: Completed flowsheets using top-$p$ sampling}
\end{figure}

In the first, rather simple example, we start with one raw material and a heat exchanger as the input sequence. The start sequence represented as an SFILES~2.0 string is \texttt{(raw)(hex)}. First, we proceed with beam search as the decoding strategy to complete the SFILES~2.0 string (blue is the generated text). We take the three sequences with the highest sequence probabilities (according to Equation~\eqref{eq:final_cond_prob}. Second, we generate three examples, using top-$p$ sampling to complete the same input sequence. 
\begin{lstlisting}
Start: (raw)(hex)
Beam search (descending sequence probability): 
1. (raw)(hex)@(r)[(prod)](hex)(flash)[{tout}(prod)]{bout}(prod)@
2. (raw)(hex)@(r)(hex)(dist)[{tout}(hex)(prod)]{bout}(hex)(prod)@
3. (raw)(hex)@(r)(hex)(flash)[{tout}(hex)(prod)]{bout}(hex)(prod)@
Top-p-sampling:
1. (raw)(hex)@(mix)<1<&|(raw)&|(dist)[{tout}(prod)]{bout}(dist){bout}1{tout}(prod)@
2. (raw)(hex)@(r)(hex)(dist)[{bout}(hex)(prod)]{tout}(hex)(prod)@
3. (raw)(hex)@(dist)[{tout}(prod)]{bout}(hex)(prod)@
\end{lstlisting}
The sequences generated with beam search are grammatically correct SFILES~2.0 strings that can be converted to the flowsheet graphs shown in Figures~\ref{fig:Ex1b1}, \ref{fig:Ex1b2}, and~\ref{fig:Ex1b3}. 
The three generated sequences with top-$p$ sampling are also grammatically correct and can be converted to flowsheet graphs, shown in Figures~\ref{fig:Ex1p1}, \ref{fig:Ex1p2}, and~\ref{fig:Ex1p3}.
The order of the generated sequences with top-$p$ sampling is random and does not indicate an order of the sequence probabilities.

The example aims to show that the Generative Flowsheet Transformer is able to build a realistic flowsheet starting from an almost empty flowsheet graph. Beam search maximizes the final probability of the sequence, so in this case the generated flowsheets in Figure~\ref{fig:BeamEx1complete} result in very short flowsheets with unit operation sequences (reactor, heat exchanger and flash/distillation system) commonly found in many chemical processes. Using top-$p$ sampling can lead to a lower final probability of the generated sequences, thus, less common topologies such as in Figure~\ref{fig:Ex1p1}. Nevertheless, both selected decoding strategies lead to correct but relatively simple final flowsheets, engineers can quickly design in practice. A more realistic scenario would be the autocompletion of more complex flowsheets (e.g. in Section~\ref{sec:completion1}) or an interactive approach between the Generative Flowsheet Transformer and a flowsheet designer. 

\subsection{Example 2}\label{sec:completion2} In the second example, we consider a more complex input sequence that contains a mixer and reactor, with the outlet of the reactor being the bottom inlet of an absorption column. The goal of this example is to study if the autocompletion model can reference a recycle to the mixer in the generated tokens and, second, complete the three other streams associated with an absorption column, in specific, a top inlet, bottom outlet, and top outlet. 

\begin{lstlisting}
Start: (raw)(mix)<1(r){bin}(abs)
Beam search (descending sequence probability): 
1. (raw)(mix)<1(r){bin}(abs)@<&|(raw)={tin}=&|[={tout}=(prod)]={bout}=(dist)[{tout}(prod)]{bout}(dist){tout}=1={bout}(prod)@
2. (raw)(mix)<1(r){bin}(abs)@<&|(raw)={tin}=&|[={bout}=(prod)]={tout}=(dist)[{bout}(prod)]{tout}(hex)=1=@
3. (raw)(mix)<1(r){bin}(abs)@<&|(raw)={tin}=&|[={tout}=(prod)]={bout}=(dist)[{tout}(prod)]{bout}(dist)[{tout}(prod)]{bout}(dist){tout}=1={bout}(prod)@
Top-p:
1. (raw)(mix)<1(r){bin}(abs)@<&|(raw)={tin}=&|[={bout}=(prod)]={tout}=(dist)[{tout}(prod)]{bout}(pp)(hex)=1=@
2. (raw)(mix)<1(r){bin}(abs)@[{bout}(prod)]{tout}(dist)[{bout}(prod)]{tout}(mix)<&|(raw)&|(pp){tin}1@
3. (raw)(mix)<1(r){bin}(abs)@<&|(raw)={tin}=&|[={bout}=(prod)]={tout}=(pp)(hex){1}(dist)[{tout}(hex){1}(prod)]{bout}(hex)(comp)=1=@
\end{lstlisting}

\begin{figure}[H]
	\centering
	\begin{subfigure}{\textwidth}
      \centering
      \includegraphics[scale=0.5]{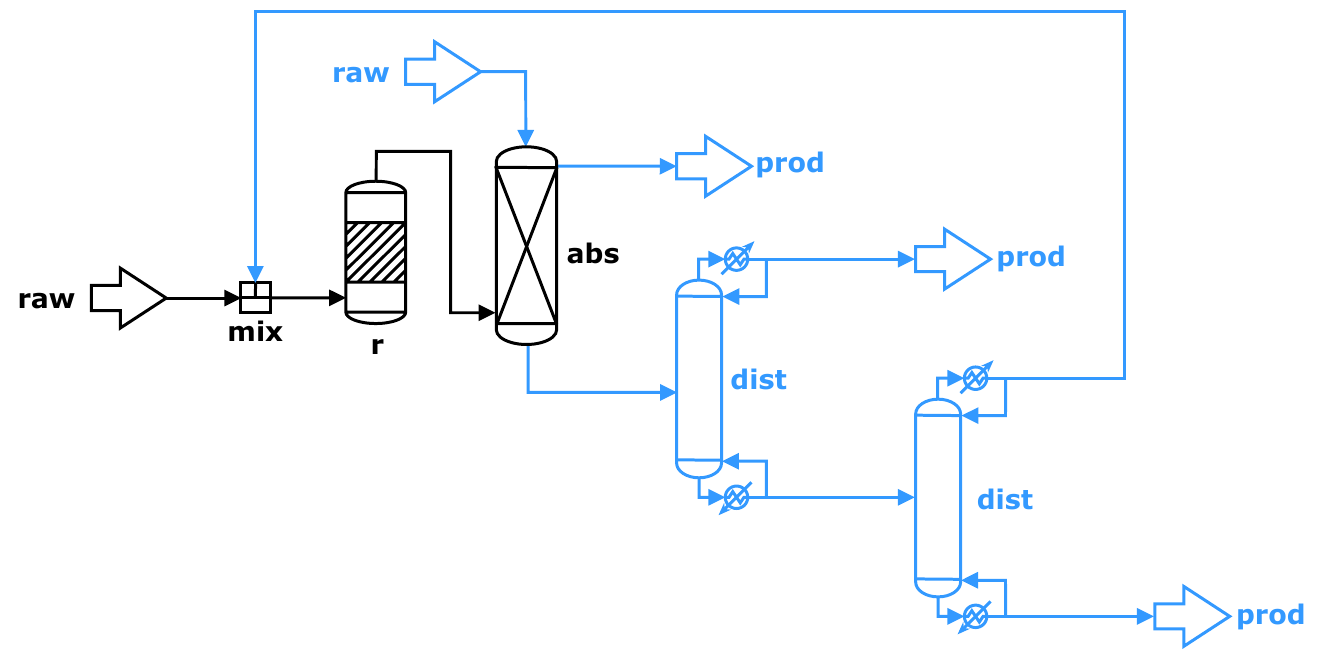}
      \caption{1. generated sequence}
      \label{fig:Ex2b1}
    \end{subfigure}
   \par\bigskip 
    \begin{subfigure}{\textwidth}
       \centering
       \includegraphics[scale=0.5]{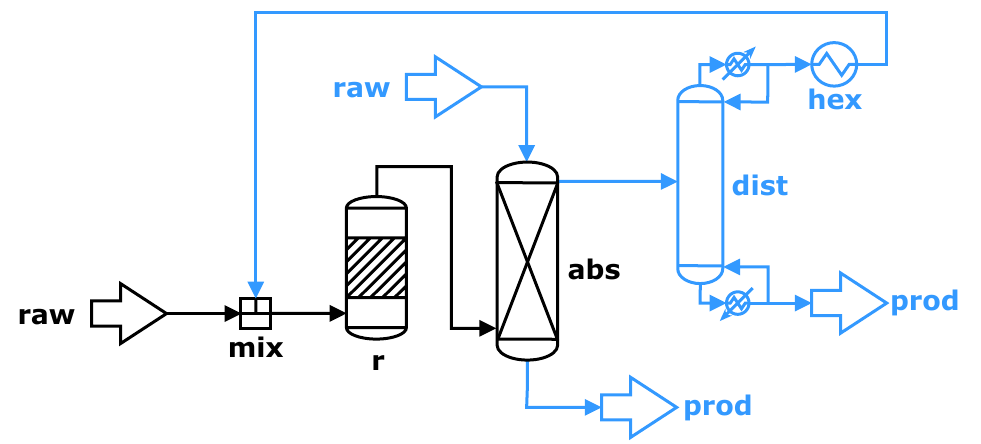}
       \caption{2. generated sequence}
       \label{fig:Ex2b2}
    \end{subfigure}
     \par\bigskip 
    \begin{subfigure}{\textwidth}
    \centering
       \includegraphics[scale=0.5]{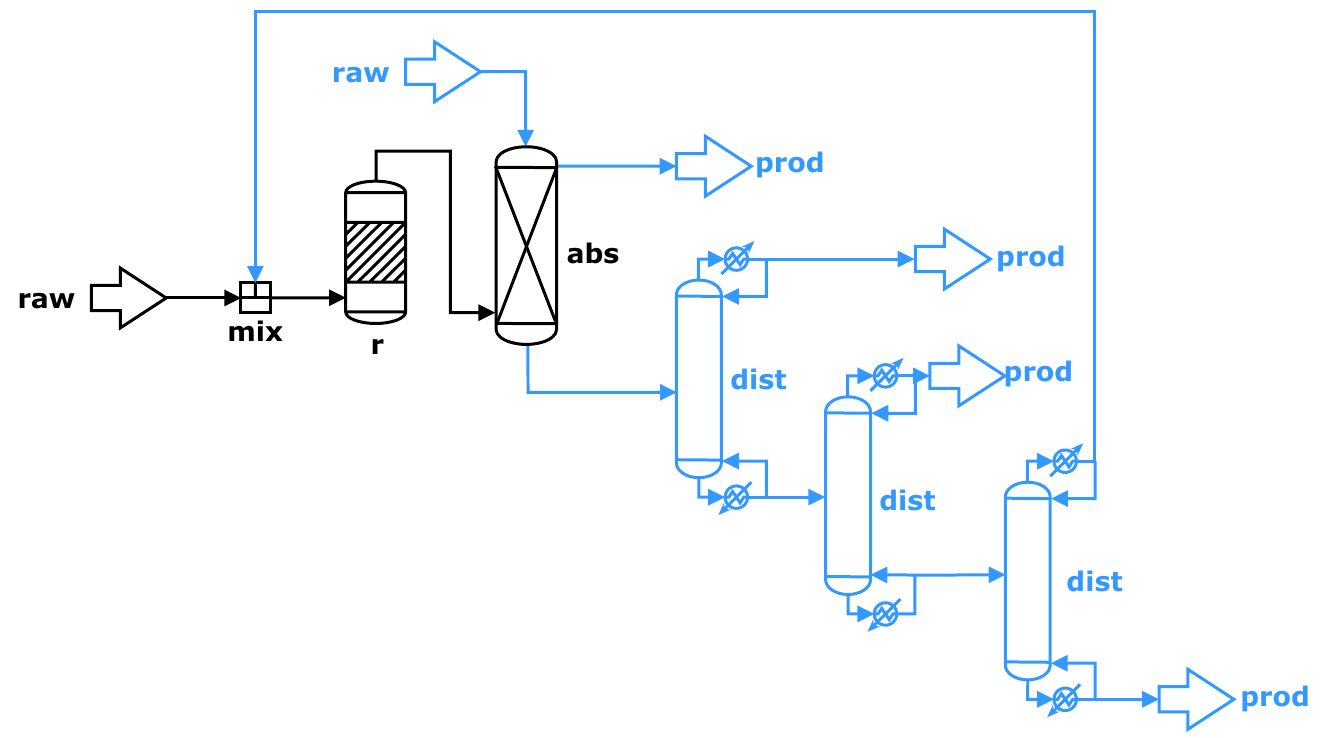}
       \caption{3. generated sequence}
       \label{fig:Ex2b3}
    \end{subfigure}
	\caption{Example 2: Completed flowsheets using beam search}
	\label{fig:BeamEx2complete}
\end{figure}

Again in this example the beam search outputs lead to grammatically correct SFILES~2.0 strings. The result demonstrates the model's ability beyond predicting the next unit operation in a flowsheet. It is able to close the recycle stream by inserting a \texttt{1} and generate all streams associated with an absorption column including a new raw material stream, as shown in Figures~\ref{fig:Ex2b1} to~\ref{fig:Ex2b3}. Similar to the previous example, beam search maximizes the final sequence probability which leads to adding only few common unit operations (only distillation columns and one heat exchanger) while completing the recycle and absorption column streams.

\begin{figure}[H]
\centering
\begin{subfigure}{\textwidth}
  \centering
  \includegraphics[scale=0.5]{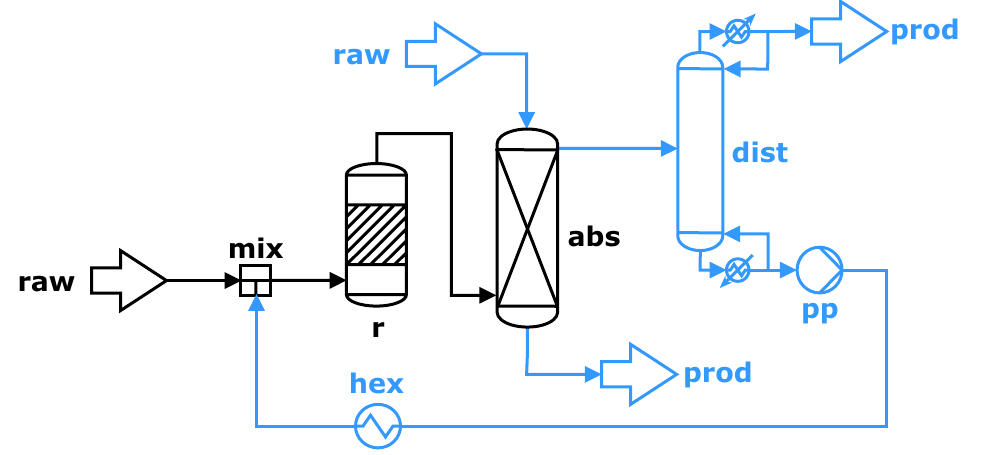}
  \caption{1. generated sequence}
  \label{fig:Ex2p1}
\end{subfigure}
\par\bigskip 
\begin{subfigure}{\textwidth}
   \centering
   \includegraphics[scale=0.5]{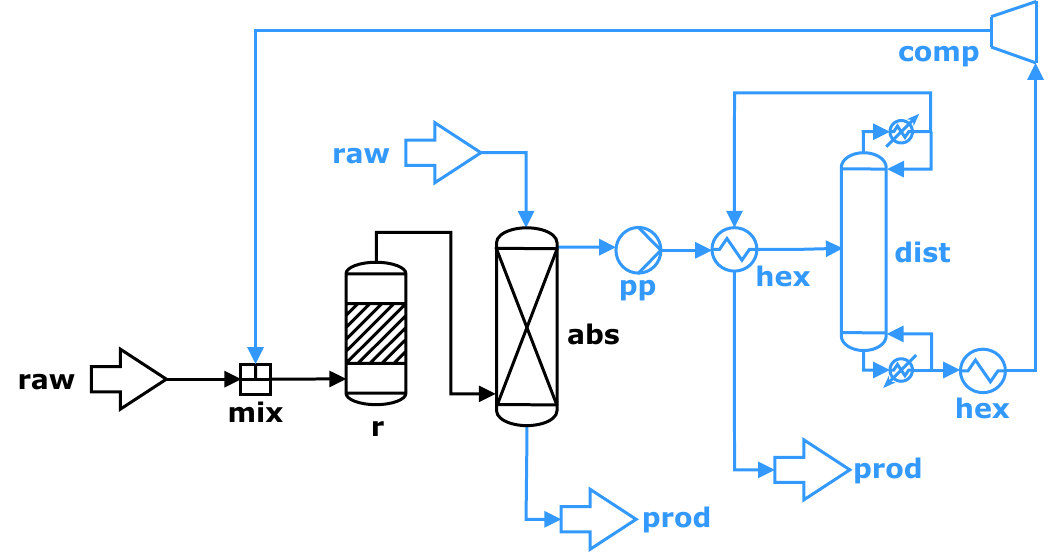}
   \caption{3. generated sequence}
   \label{fig:Ex2p3}
\end{subfigure}
\caption{Example 2: Completed flowsheets using top-$p$ sampling}
\end{figure}

Flowsheet completion with top-$p$ sampling as decoding strategy only leads to two out of three grammatically correct SFILES~2.0 strings, i.e., examples 1 and 3, shown in Figures~\ref{fig:Ex2p1} and~\ref{fig:Ex2p3}. As expected, the generated sequences comprise unit operations that occur less often in the training data, such as pumps or compressors. Moreover, the flowsheet in Figure~\ref{fig:Ex2p3} shows some logical errors. For example, a pump is applied to the outlet of the absorber and a compressor is applied to the bottom outlet of the distillation column. 

\subsection{Current limitations and future directions}
\label{sec:limitations}
The training results in Section~\ref{sec:results_training} and the autocompletion examples in Subsections~\ref{sec:completion1} and~\ref{sec:completion2} demonstrate for the first time, that a NLP model can learn to autocomplete chemical process topologies. 
In particular, the model is capable of learning typical topological patterns from the flowsheet data.  
However, the examples also demonstrate a number of current limitations that require future research. 
In example 1, a number of flowsheets are predicted based on a short given sequence of a raw material stream and a heat exchanger. 
While the predicted flowsheets use common patterns of unit operations, they are not yet meaningful for practical applications.
Currently, the NLP model takes only topologies of flowsheets as inputs and has no information about the context (e.g., concentrations, materials, operating points). However, this information is essential to fully describe a process.
Moreover, the model is not able to modify the input sequence. This flexibility is important, for instance, to perform heat integration of already used heat-exchangers or insert unit operations (e.g. mixer) at the beginning of the process. 
In example 2, the algorithm also suggests some sequences of unit operations that are physically wrong. 
For instance, a compressor is added to the bottom outlet of a distillation column.

We identify three promising directions for future research that have the potential to overcome the current limitations of our proposed approach:
\begin{enumerate}
    \item Encode additional information about the process, such as chemical compounds, reactions, stream compositions, stream conditions (temperatures and pressures), and operating points of unit operations. This could be achieved in two ways: 
    \begin{itemize}
        \item Extension of SFILES~2.0 language to encode additional information,
        \item Modification of the model architecture to include additional process information as numerical input features, e.g., in the head of the language model. Also our previous work shows that graph-based machine learning models can process additional flowsheet information~\cite{Stops2022}. 
    \end{itemize}
    \item Increase the quantity and quality of training data. This can be achieved in multiple ways, e.g.: 
    \begin{itemize}
        \item Collect more flowsheet data including the necessary additional process information from literature and patents~\cite{Schweidtmann2022Flowsheetmining,Theisen-2022-Digitization},
        \item Collect and extract process information from simulation files,
        \item Collect data directly in a simulation software. 
    \end{itemize}    
    \item Integrate physical knowledge and rules into the machine learning approach~\cite{Venkatasubramanian2018,Schweidtmann2021}. This could reduce the data demand and enhance prediction quality. 
\end{enumerate}
Finally, we suggest two explorative directions for future research. First, we suggest to extend the model architecture such that that input sequence can be adapted. This would allow to account for heat integration with previous process units. Second, we suggest to operate on a phenomena level instead of a unit operation level~\cite{lutze2013phenomena}. This could potentially facilitate learning. Also, it would allow to account for intensified unit operations.
Overall, our results demonstrate that the proposed transformer model can successfully learn from flowsheet topologies. 
Future research in the aforementioned areas has the potential to further increase the Technology Readiness Level (TRL) and pave the way for for industrial applications.

\section{Conclusions}
We propose a novel method to learn from chemical process flowsheets and provide flowsheet structure recommendations for engineers performing process synthesis. We created two data sets, the first one consisting of synthetically generated and the second one consisting of real flowsheets in graph format. 
Using the conversion algorithm for the automated conversion between flowsheet graphs and SFILES~2.0 strings~\cite{GabrielVogel}, we automatically generated the corresponding text-based SFILES~2.0 data sets.
We trained a generative transformer language model on over 8,000 flowsheet topologies. 
The trained generative transformer model shows the ability to learn the grammatical structure of the SFILES~2.0 language and the patterns contained in the flowsheet topologies. 
Consequently, the results demonstrate that using the trained model for causal language modeling is a promising strategy to autocomplete flowsheet topologies.
However, we also identified current limitations regarding applicability of the model in realistic process design scenarios. 

We identify three main directions for future research: First, future work should focus on creating a larger data set of real flowsheets. To achieve this, we investigate the mining of flowsheets from literature~\cite{Schweidtmann2022Flowsheetmining} and the digitization of PFDs and P\&IDs~\cite{Theisen-2022-Digitization}. 
Also, we strive for cooperation with industrial partners to work on industrial flowsheets.
Second, including the context of the chemical processes is indispensable for improving the significance and applicability of the model’s predictions. This includes, for instance, chemical property data, the reaction and conversion of substances, stream compositions, and operating conditions. 
Third, building hybrid models by combining our data-driven approach with process design knowledge might contribute to more intelligent systems~\cite{Schweidtmann2021}. 
Addressing these open research challenges will pave the way for the application of the proposed autocompletion method in industry.
Ultimately, we envision that autocompletion of flowsheets and operating point recommendation will become a standard tool in future process simulation software.  

\section*{Acknowledgements}
This publication is part of the project “ChemEng KG – The Chemical Engineering Knowledge Graph” with project number 203.001.107 of the research programme “Open Science (OS) Fund 2020/2021” which is (partly) financed by the Dutch Research Council (NWO). GV acknowledges the ERASMUS Plus scholarship for his reserach stay at the Process Intelligence Reserach group. The authors acknowledge the fruitful discussions with Prof. R. Gani and Prof. J. Grievink on the combination of data-driven and mechanistic models.

\bibliographystyle{elsarticle-num}

\end{document}